\documentclass[letterpaper]{article} 
\usepackage{aaai24}  
\usepackage{times}  
\usepackage{helvet}  
\usepackage{courier}  
\usepackage[hyphens]{url}  
\usepackage{graphicx} 
\usepackage{natbib}  
\usepackage{caption} 
\usepackage{algorithm}

\usepackage{algorithmicx}

\usepackage{algpseudocode}
\usepackage{multirow}
\usepackage{amsmath}
\usepackage{textcomp}
\usepackage[table,xcdraw]{xcolor}
\usepackage{appendix}

\usepackage{makecell}

\usepackage{newfloat}
\usepackage{listings}
\DeclareCaptionStyle{ruled}{labelfont=normalfont,labelsep=colon,strut=off} 
\floatstyle{ruled}
\newfloat{listing}{tb}{lst}{}
\floatname{listing}{Listing}
\title{Generalized Planning for the Abstraction and Reasoning Corpus}

\author {
    Chao Lei, Nir Lipovetzky, Krista A. Ehinger \\
}
\affiliations {
    School of Computing and Information Systems, The University of Melbourne, Australia\\
    clei1@student.unimelb.edu.au, \{nir.lipovetzky, kris.ehinger\}@unimelb.edu.au

}

\usepackage{bibentry}

\begin{document}

\maketitle

\begin{abstract}
The Abstraction and Reasoning Corpus (ARC) is a general artificial intelligence benchmark that poses difficulties for pure machine learning methods due to its requirement for fluid intelligence with a focus on reasoning and abstraction. In this work, we introduce an ARC solver, Generalized Planning for Abstract Reasoning (GPAR). It casts an ARC problem as a generalized planning (GP) problem, where a solution is formalized as a \textit{planning program} with \textit{pointers}. We express each ARC problem using the standard Planning Domain Definition Language (PDDL) coupled with \textit{external functions} representing object-centric \textit{abstractions}. We show how to scale up GP solvers via domain knowledge specific to ARC   in the form of  restrictions over the actions model, predicates, arguments and valid structure of planning programs. Our experiments demonstrate that  GPAR outperforms the state-of-the-art solvers on the object-centric tasks of the ARC, showing the effectiveness of GP and the expressiveness of PDDL to model ARC problems. The challenges provided by the ARC benchmark motivate research to advance existing GP solvers and understand new relations with other planning computational models. Code is available at \url{github.com/you68681/GPAR}.

\end{abstract}

\section{Introduction}

\begin{figure}[ht]
 
\centering
\includegraphics[scale=0.222]{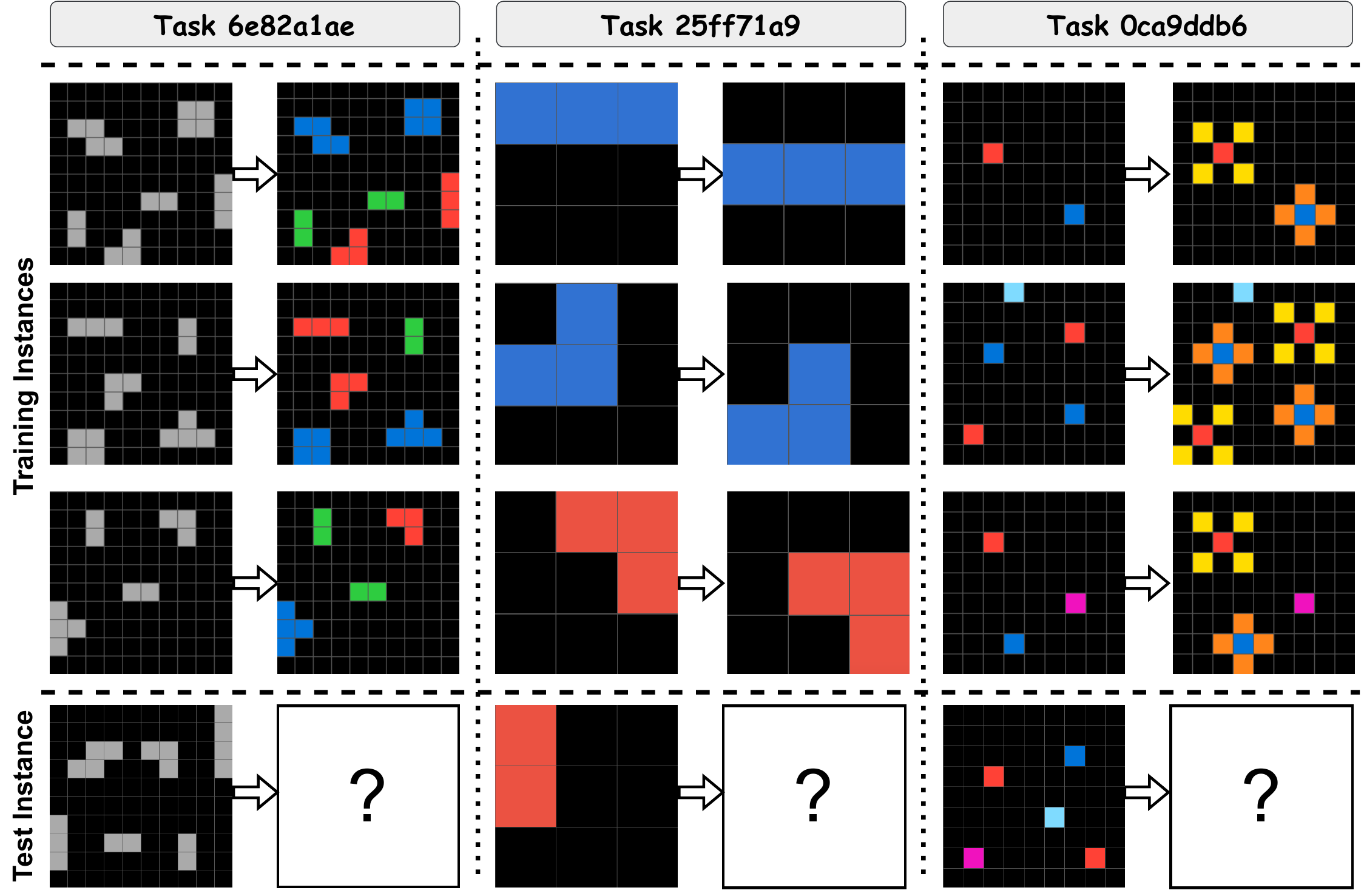}        
\caption{Three example tasks from the ARC. For a given task, each row contains an input-output image pair as a training instance, and the bottom row is the test instance. The goal of the solver is to learn from the training instances how to generate the output for the test instance.}
\label{fig1}
 
\end{figure}
Abstract visual reasoning tasks have been used to understand and measure machine intelligence \citep{malkinski2022review,barrett2018measuring,moskvichev2023conceptarc}. One of these tasks, the Abstraction and Reasoning Corpus (ARC) introduced by \citet{chollet2019measure}, remains an open challenge. ARC tasks are challenging for machines because they require object recognition, abstract reasoning, and procedural analogies \citep{johnson2021fast,acquaviva2021communicating}.
ARC comprises 1000 unique tasks where each task consists of a small set (typically three) of input-output image pairs for training, and generally one or occasionally multiple test pairs for evaluation (Figure \ref{fig1}). Each image is a 2D grid of pixels with 10 possible colors. ARC tasks require inferring the underlying rules or procedures from a few examples based on \textit{core knowledge} priors including  objectness, goal-directedness, numbers and counting, topology and geometry. 

\citet{chollet2019measure} suggested a hypothetical ARC solver that includes a program synthesis engine for candidate solutions generation within a ``human-like reasoning Domain Specific Language (DSL)". Few successful solvers have followed this approach. Inspired by human strategies for solving ARC tasks \citep{johnson2021fast,acquaviva2021communicating}, \citet{xu2022graphs} proposed an object-centric approach, Abstract Reasoning with Graph Abstractions (ARGA) that adopts a graph-based DSL representation and performs a constraint-guided search to find programs in the DSL solving the task. ARGA demonstrates considerable generalization ability and efficient search. However, due to the limited expressiveness of the DSL, its performance is worse for some ARC tasks than the Kaggle competition’s first-place solution \cite{Kaggle2020}. This algorithm searches in a directed acyclic graph to synthesize program solutions over a hand-crafted DSL, where each search node represents an image transformation applied to its parent node.

\textit{Generalized planning} (GP), a program synthesis approach that studies the representation and generation of solutions that are valid for a set of problems, is well suited to the ARC \citep{srivastava2008learning, hu2011generalized, jimenez2019review}.  Solutions, known as generalized plans, can be formalized as \textit{planning programs} with \textit{pointers} \cite{segovia2019computing} where conditional statements, and looping and branching structures allow the compact representation of solutions. Recent advances in GP solvers have significantly improved the search efficiency, enabling the applicability of GP over new challenging benchmarks \cite{lei2023novelty}.

In this work, we propose an ARC solver called Generalized Planning for Abstract Reasoning (GPAR), which models each ARC task as a generalized planning problem and adopts a state-of-the-art planner to perform program synthesis. We improve existing graph abstractions to promote greater object awareness and introduce a novel DSL based on the Planning Domain Definition Language (PDDL)~\cite{haslum2019introduction}, where hybrid declarative and imperative modeling languages are combined to guarantee enough expressivity, and  represent the transition function concisely. Our main contributions are:
1) a novel method to solve abstract reasoning tasks based on generalized planning, which achieves the state-of-the-art performance over the ARC benchmark; 2) an encoding based on PDDL which enables the adoption of alternative planning models for visual reasoning; 3) the usage of novel ARC domain knowledge that other ARC solvers can use to reduce the size of the solution space.

\section{Background}

\subsubsection{Planning Domain Definition Language}
PDDL, the \textit{de facto} standard modeling language for several different classes of planning problems, allows the usage of automated planning solvers to find plans that map an initial state into one of the goal states of a transition system~\cite{haslum2019introduction}. PDDL divides the representation of a planning problem into two parts, a \textit{domain} $\mathcal{D}$ to define the \textit{predicates} and \textit{action schemes}, consisting of  preconditions and effects, whose parameters can be instantiated with a typed-system of constant objects, and a \textit{problem} or \textit{instance}  $\mathcal{I}$, defining the \textit{objects}, \textit{initial state}, and \textit{goal formula} that entails a set of goal states. Different problems with the same domain can be created by changing any problem definition element: objects, initial state or goal conditions. 

The induced transition system can be traversed through the application of actions. In fact, a plan is typically composed of a sequence of actions. To be applicable, the action preconditions need to be true in a state, and the resulting state is generated by incorporating the effects of the action, where some ground atoms of a predicate become true or false. Preconditions and effects are generally described through formulas in first-order logic. It is known that certain effects can be described more concisely by alternative languages or simulators, better equipped to reason over complex numeric operations, for example~\citet{dornhege2012semantic}. We use PDDL with \textit{external functions} whose denotation is specified using imperative languages to express complex preconditions and effects of the ARC tasks~\cite{frances2017purely}.

\begin{figure}[ht]
\centering
\includegraphics[scale=0.32]{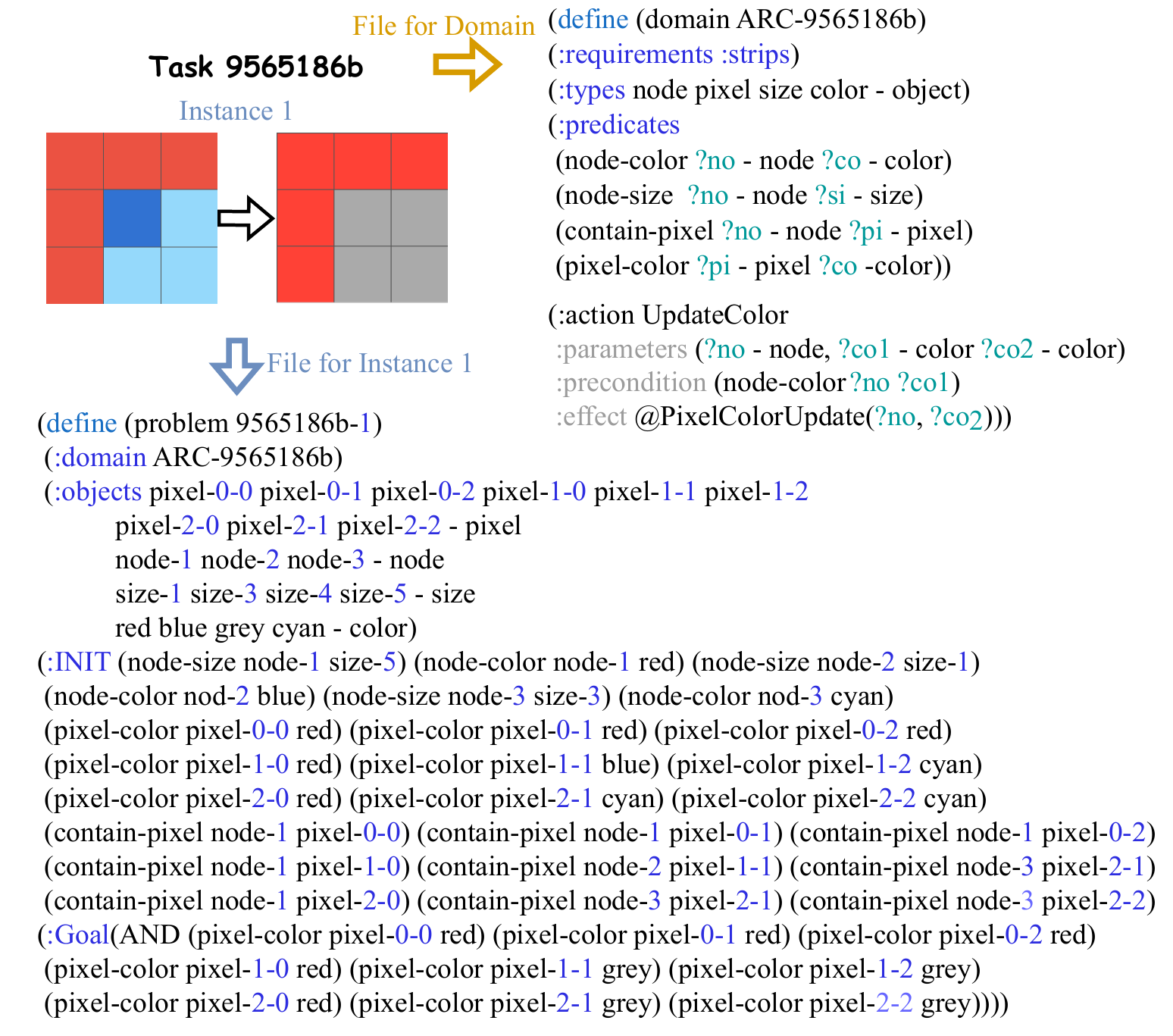}      
\caption{A PDDL example for a fragment of an ARC task. }
\label{fig2} 
\end{figure}

Figure \ref{fig2} presents a PDDL domain and instance file for a fragment of an ARC task. Parameters of action schemes and predicates are preceded by the ``?'' symbol, and external functions are preceded by the ``\textit{@}'' symbol.

\subsubsection{Generalized Planning} GP aims to solve a finite set of classical planning problems $\mathcal{P}$ over the same domain $\mathcal{D}$, where each instance $\mathcal{I}$ may differ in the initial state $I$,  goal conditions $G$, or objects $\Delta$. A GP solution is a single \textit{program} that produces a valid plan for every classical planning instance.

\subsubsection{Planning Programs with Pointers} \textit{planning programs} with \textit{pointers} $Z$, where each pointer indexes a type of object in $\mathcal{P}$, compactly describe a scalable solution space for GP \citep{segovia2022scaling}. A planning program $\Pi$, is a sequence of programmable instructions, i.e. $\Pi = \langle {w_0},\ldots, {w_{n-1}} \rangle$, with a given maximum number of program lines $n$. 

\begin{figure}[!ht]
 
\centering
\includegraphics[scale=0.20]{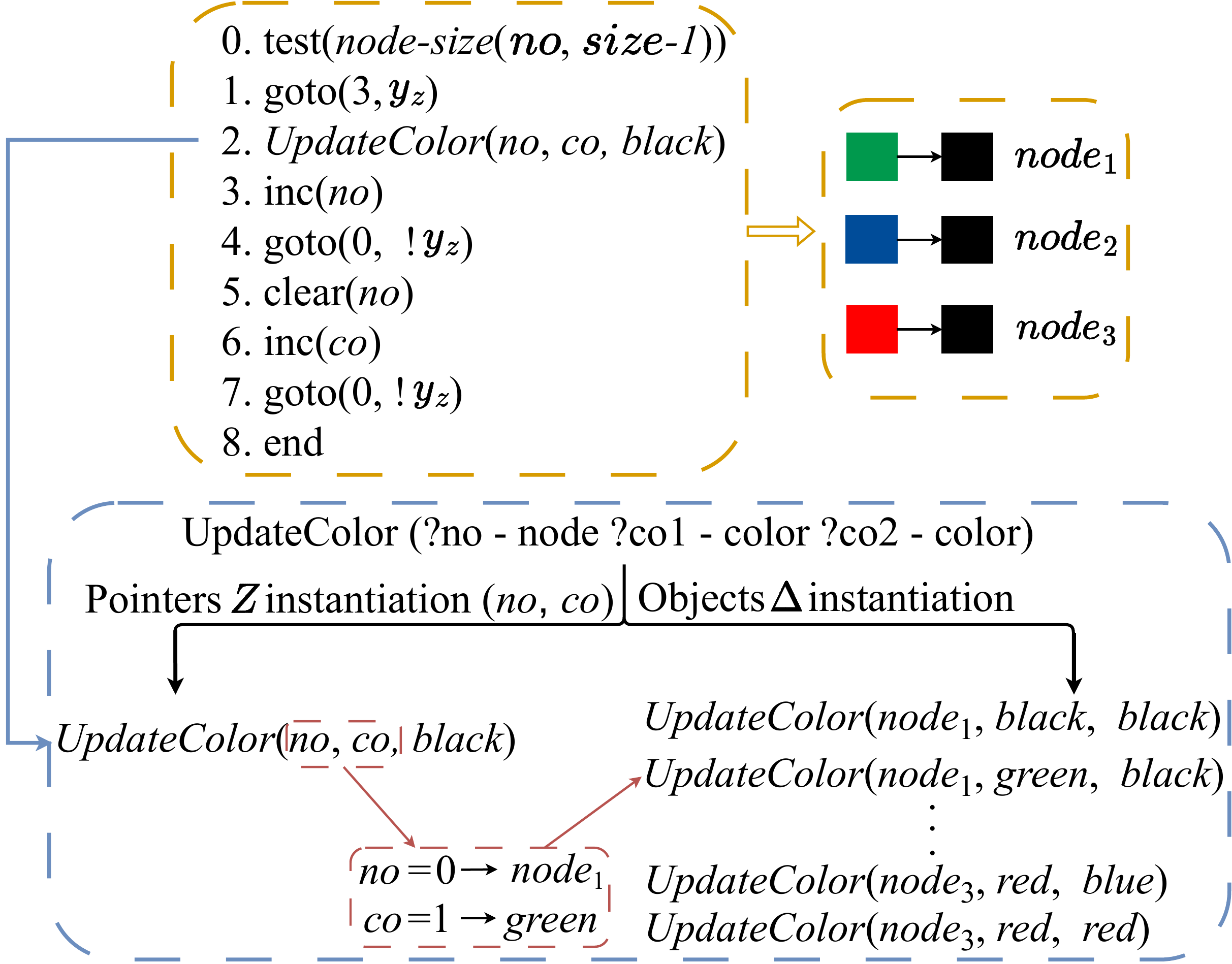}
\caption{A planning program $\Pi$ to alter size-1 nodes with different colors to black.  Pointers $no$ indexes node objects $\{\textit{node$_1$}, \textit{node$_2$}, \textit{node$_3$}\}$, and  $co$ indexes color objects $\{\textit{black}, \textit{green}, \textit{blue}, \textit{red}\}$. 
}
\label{fig3} 
\end{figure}

An instruction $w_i$, where $i$ is the location of the \textit{program line}, $0\leq i < n-1$, can either be a planning action $a_z$ instantiated from the action scheme over pointers or constant objects, a RAM action $a_r$ to manipulate pointers, a \texttt{test} action to return the interpretation of a predicate, a \texttt{goto} instruction for non-sequential execution, or a special \texttt{end} instruction for termination that is always programmed in the last line. A \texttt{goto} instruction is a tuple $\textrm{goto}$($i^{\prime}$, $y$), where $i^{\prime}$ is the destination line, and $y$ is a proposition that captures the result of the last execution of RAM or \texttt{test} action. We refer the reader to~\citet{segovia2019computing} for a full specification.

When $\Pi$ begins to execute on an instance $\mathcal{I}_t$, a \textit{program state} pair $(s, i)$  is initialized to $(I_t,0)$, where $I_t$ is the initial state of instance $\mathcal{I}_t$. Meanwhile, pointers are equal to zero, and all of $y$ is set to \textit{False}. An instruction $w_i \in \Pi$ updates $(s,i)$ to ($s^{\prime}$, $i+1$) when $w_i=a_z$ or $w_i=a_r$, where $s^{\prime}$ is the resulting state if $w_i$ is applicable, or, $s^{\prime} = s$ otherwise. An instruction relocates the program state to $(s, i^{\prime})$ when ${w_{i}} = \textup{go}$($i^{\prime}$, $y$) if $y$ holds in $s$, or to the next line otherwise $(s, i + 1)$. $\Pi$ is a solution for $\mathcal{I}_t$ if $\Pi$ terminates in ${(s,i)}$ and meets the goal condition, i.e. ${w_{i}}=\texttt{end}$ and  $G \subseteq s$. $\Pi$ is a solution for the GP problem $\mathcal{P}$, iff $\Pi$ is a solution for every instance $\mathcal{I}_t\in\mathcal{P}$.

The upper part of Figure \ref{fig3} illustrates a planning program discovered by our solver that updates the color of any size-1 node (a collection of pixels) to black using two pointers, $no$ and $co$, to iterate over node and color objects. The bottom part illustrates how a single planning action can represent a large set of object-instantiated action executions. The inner loop, lines 0 to 4, updates a size-1 node $no$ with color $co$ to black. If the precondition of the action \textit{UpdateColor} is false (see Fig~\ref{fig2}), then the effect of the action will not be executed. When $no$ points to the last node object, line 3 fails to increment the pointer, and lines 5-7 are executed to set $no$ to the first node, and let $co$ point to the next color. When all colors have been tried, then the program will end.

\section{Abstraction over ARC}

Abstraction enables object awareness in GPAR to allow actions to modify a group of pixels at once rather than individually, resulting in a smaller search space. Object cohesion is central to human visual understanding \cite{spelke2007core}, and humans doing ARC tasks seem to come up with solutions that involve objects and object relations \cite{acquaviva2021communicating, johnson2021fast}. 
However, part of the challenge of the ARC tasks comes from the fact that there are multiple ways to interpret the images, and different tasks may require different ``objects''. Therefore, we consider multiple possible abstract representations. As in \citet{xu2022graphs}, we represent an image as a graph of \textit{nodes} representing objects and their spatial relations. 

\begin{figure}[t]
 
\centering
\includegraphics[scale=0.6]{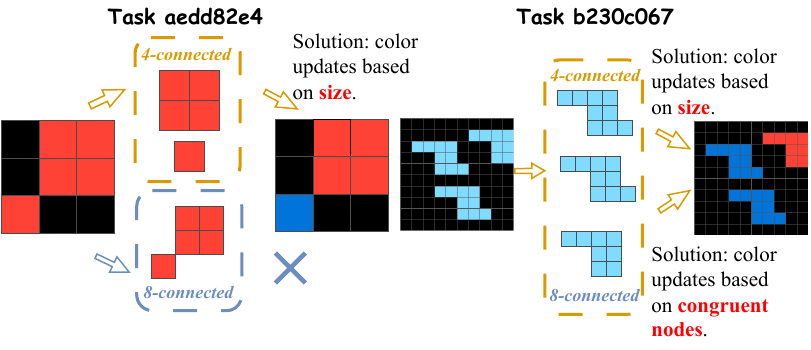}

\caption{An example task for which the 4- vs. 8-connected abstractions produce different nodes (left), and an example task where identified nodes are the same (right).}
\label{fig4} 
\end{figure}

Inspired by \citet{xu2022graphs}, we consider the following abstractions: 1) \textit{4-connected}, which treats 4-connected components as nodes, excluding the background; 2) \textit{8-connected}, which treats 8-connected components as nodes, excluding the background; 3) \textit{same-color}, which treats all pixels of the same color as a node, regardless of their connectivity; 4) \textit{multi-color}, which treats all non-background colors as the same for the purposes of forming 4-connected and 8-connected components (thus allowing the creation of multi-colored nodes); 5) \textit{vertical} and \textit{horizontal}, which form nodes of columns or rows, respectively, of same-colored non-background pixels; 6) \textit{pixels}, which treats each pixel as a node; 7) \textit{image}, which treats the entire image as a single node; 8) \textit{max-rectangle}, which recognizes the maximum rectangle that can be inscribed within a 4-connected component as a node and subsequently processes the remaining pixels as 4-connected components, in both non-background and background regions.

Different node definitions can compensate for the limitations of a certain abstraction, such as in Figure \ref{fig4} (left), where only the 4-connected abstraction is reasonable. However, two abstractions may produce identical nodes for a given ARC task, as in Figure \ref{fig4} (right). To avoid duplication, we only consider an abstraction if it produces a different node representation in terms of size, color, or shape for at least one training instance.

\subsubsection{Node Attributes.}
Each identified node is associated with basic attributes, including color, size, and shape. Shape can be a \textit{single pixel}, \textit{square}, \textit{rectangle},  \textit{horizontal line}, \textit{vertical line}, \textit{left diagonal line}, \textit{right diagonal line} or \textit{unknown}. To address tasks involving counting or sorting objects, nodes with the largest and smallest size, odd and even size, and most and least frequently occurring color are also indicated. 

For some abstractions, the aforementioned attributes are inappropriate, and alternative attributes are used. For multi-color nodes, the color attribute is omitted. When considering either pixel nodes or image nodes, only the most and least frequent color are identified. For pixel nodes, we use additional attributes to represent image geometry, denoting which nodes are on the image borders, centric-diagonal, middle-vertical and middle-horizontal lines and to detect and remove pixels that are potentially noise (defined as 4-connected components with a size of 1 pixel).

\subsubsection{Relations between Nodes}
We define three types of node relations, \textit{spatial}, \textit{congruent}, and \textit{inclusive}, applicable to all node definitions except for pixel and image nodes. Spatial relations, right, left, up, and down, exist between two nodes iff there is at least one pixel in each node with the same coordinate value along either axis. Diagonal spatial relations are considered for two nodes whose shapes are not \textit{unknown} and whose corners align on the same diagonal axis. Congruent relations are defined between nodes with identical shapes and sizes, and nodes with the same color. Inclusive relations specify which nodes contain or partially contain other nodes. A node contains another node if all the pixels of the contained node lie within the borders of the containing node. A node partially contains another node if the above relation holds and the borders of the contained node touch the boundaries of the image. Node attributes and relations are sourced from \textit{core knowledge} priors and extracted through standard image processing approaches.

\begin{table}[]
\captionsetup{font={small}}
\small
\setlength{\tabcolsep}{5pt}
\renewcommand{\arraystretch}{1}
\begin{tabular}{p{0.23\columnwidth}|p{0.7\columnwidth}}
{Object Types} & Possible Associated Objects                                                                                       \\ \hline\hline
{\textsc{node}}         & \textit{node-0}, \ldots, \textit{node-n}.                                                                                                    \\
{\textsc{pixel}}        & \textit{pixel-0-0}, \ldots, \textit{pixel-29-29}.                                                                                               \\
{\textsc{color}}        & \textit{color-0}, \ldots, \textit{color-9}.                                                                                                 \\
{\textsc{size}}        & \textit{size-1}, \ldots, \textit{size-9000}.                                                                                                     \\
{\textsc{step}}         & \textit{one}, \textit{max}.                                                                                                                \\
{\textsc{rotation}}     & \textit{90$^{\circ}$}, \textit{180$^{\circ}$}, \textit{270$^{\circ}$}.                                                                                                           \\
{\textsc{f-direction}}   & \textit{vertical}, \textit{horizontal},  \textit{left-diagonal}, \textit{right-diagonal}.                                                                \\
{\textsc{m-direction}}   &  \textit{left}, \textit{right}, \textit{up}, \textit{down},  \textit{left-up}, \textit{left-down}, \textit{right-up}, \textit{right-down}.           \\
{\textsc{shape}}        & \textit{single-pixel}, \textit{square}, \textit{rectangle}, \textit{vertical-line}, \textit{horizontal-line}, \textit{left-diagonal-line}, \textit{right-diagonal-line}, \textit{unknown}.
\end{tabular}
\caption{Object types and associated objects in our DSL.}
\label{table1}
\end{table}

\section{Domain-Specific Language}

PDDL leverages a subset of first-order logic, a powerful tool to represent knowledge for reasoning purposes. It provides a structured and concise way to express relations between objects and properties \cite{genesereth2012logical,levesque1986knowledge}. PDDL describes each ARC task through a single domain file and a finite set of instance files, one for each input-output image pair. The domain file contains the relations between nodes and their attributes, modeled as predicates, and node transformations, modeled by action schemes. Action schemes and predicates are instantiated through the objects specified in the instance file, where the conjunctive formula of  instantiated predicates describes the initial state representing an input image, and a goal state modeling the target image configuration.

Given an ARC task, Table \ref{table1} shows the available objects and their types, while Table \ref{table2} presents the available predicates to model node attributes and their relations. We differentiate between predicates that can be interpreted by the \texttt{test} action to condition a \texttt{goto} instruction, indicated by the \texttt{test}$_p$ column, and predicates whose main purpose is to encode knowledge in our DSL. Table \ref{ExampleAction} introduces a subset of the main action schemes included in our DSL, where the preconditions or effects are implemented by external functions, either to check the applicability of certain actions or facilitate node transformations. We encode a mix of low-level and high-level actions, where some high-level actions encode complex transformations that would otherwise require several low-level actions. This enables the solver to reason at the appropriate level of abstraction and lower the program complexity when possible. E.g.,  \textit{SwapColor} and \textit{CopyColor} can be realized by the ground action \textit{UpdateColor} with additional program logic to manipulate pointers, but this would require increasing the number of program lines encoding a solution. 

Each abstraction is associated with its respective set of actions and predicates and a full description is available in the supplementary materials. We also consider two additional abstractions to enable complicated movement, extension, and congruent node operations, where both node definitions are the same as the 4-connected abstraction. These are only tried if no solution can be found in the simpler abstractions.

\begin{table}[t]
\captionsetup{font={small}}
\small
\setlength{\tabcolsep}{1.5pt}
\renewcommand{\arraystretch}{1.3}
\begin{tabular}{c|clcccccl}
                                                 &      & \multicolumn{1}{c}{Predicates (? Parameters)}                                                                                                                                                                                                                                                                                                                                                                                                                                      &                      &                      &                      &                      &                      &  \\ \cline{1-3} \hline\hline
\multicolumn{1}{c|}{\multirow{2}{*}{\rotatebox{90}{Attributes}}} & \texttt{test$_p$} & \begin{tabular}[c]{@{}l@{}}\textit{color-most}(\textit{color}), \textit{color-least}(\textit{color}),  \\ \textit{color-max}(\textit{node}),  \textit{color-min}(\textit{node}), \textit{size-min}(\textit{node}),\\ \textit{size-max}(\textit{node}), \textit{odd}(\textit{node}), \textit{even}(\textit{node}), \\ 
                            \textit{up-border}(\textit{node}),  \textit{down-border}(\textit{node}),\\ \textit{left-border}(\textit{node}), \textit{right-border}(\textit{node}), \\ \textit{left-diagonal}(\textit{node}), \textit{right-diagonal}(\textit{node}), \\ \textit{horizontal-middle}(\textit{node}), \textit{vertical-middle}(\textit{node}),\\ \textit{node-color}(\textit{node, color}), \textit{node-shape}(\textit{node, shape}),\\ \textit{node-size}(\textit{node, size}), \textit{denoising-color}(\textit{node, color}).\end{tabular} &                      &                      &                      &                      &                      &  \\ \cline{2-3}
\multicolumn{1}{c|}{}                            &   -   & \textit{background}(\textit{color}).                                                                                                                                                                                                                                                                                                                                                                                                                                                   & \multicolumn{1}{l}{} & \multicolumn{1}{l}{} & \multicolumn{1}{l}{} & \multicolumn{1}{l}{} & \multicolumn{1}{l}{} &  \\ \cline{1-3}
\multicolumn{1}{c|}{\multirow{2}{*}{\rotatebox{90}{Relations}}}  & \texttt{test$_p$} & \begin{tabular}[c]{@{}l@{}}\textit{node-diagonal}(\textit{node, node}),  \textit{same-color}(\textit{node, node}),  \\ \textit{congruent}(\textit{node, node}), \textit{contain-node}(\textit{node, node}),  \\ \textit{partially-contain-node}(\textit{node, node}), \\ \textit{relative-position}(\textit{node, node, m-direction}).\end{tabular}                                                                                                                                                                                                                                      &                      &                      &                      &                      &                      &  \\ \cline{2-3}
\multicolumn{1}{c|}{}                            &   -   & \textit{node-spatial}(\textit{node, node, m-direction}).                                                                                                                                                                                                                                                                                                                                                                                                                                & \multicolumn{1}{l}{} & \multicolumn{1}{l}{} & \multicolumn{1}{l}{} & \multicolumn{1}{l}{} & \multicolumn{1}{l}{} &  \\ \cline{1-3}
\multicolumn{1}{c|}{Pixel}                       & - & \textit{pixel-color}(\textit{pixel}, \textit{color}), \textit{contain-pixel}(\textit{node, pixel}).                                                                                                                                                                                                                                                                                                                                                                                                               &                      &                      &                      &                      &                      & 
\end{tabular}
\caption{Predicates in our DSL. \texttt{test$_p$} indicates following predicates can be interpreted by the \texttt{test} action; the symbol ``- ''denotes predicates can not be interpreted.}
\label{table2}
\end{table}

\subsubsection{Action Pruning}
Abstractions can introduce irrelevant actions in a domain. E.g., for the first task in Figure \ref{fig1}, actions that involve changing node positions should not be included, and in the second task, actions related to color updates should be avoided.
A similar idea is discussed by \citet{xu2022graphs}, where a newly generated node will be pruned while searching if it fails to satisfy a set of constraints generated by comparing the nodes defined by each abstraction. GPAR supports all their constraints. However, we acquire and use action constraints to prune irrelevant action schemes when generating the domain file instead of pruning generated nodes.

We consider mainly three constraints based on whether all nodes' positions, colors, or sizes remain unchanged across training input and output images. If any of the properties above hold true in the training sets, then the related actions involving  movement, color, or size updating will be pruned. Some additional constraints are included to prune actions not applicable to a given abstraction; e.g., \textit{InsertNode} is avoided when no consistent pattern (nodes with the same color, size, and shape) exists among input images. See the supplementary materials for the full list of actions associated with each constraint.

\begin{table}[t]
\captionsetup{font={small}}
\centering
\small
\setlength{\tabcolsep}{2pt}
\renewcommand{\arraystretch}{1}
\begin{tabular}{p{0.51\columnwidth}p{0.45\columnwidth}}

Action Schemes (? Parameters)                                                                                                                                                                                   & Effects                                                                                                                                                              \\ \hline \hline
\textit{UpdateColor}(\textit{node color$_1$ color$_2$})                                                                                                                           & Change the node color from color$_1$ to color$_2$.                                                                                                                          \\
\textit{SwapColor}(\textit{node$_1$ node$_2$})                                                                                                                   & Swap colors of node$_1$ and node$_2$.                                                                                                                                    \\ 

\textit{CopyColor}(\textit{node$_1$ node$_2$})                                                                                                                   & Copy the color of node$_1$ to node$_2$.                                                                                                                                    \\ 

\textit{MoveNode}(\textit{node$_1$ node$_2$})                                                                                                                   & Move node$_1$ to the boundary of node$_2$.                                                                                                                                    \\ 
\textit{MoveNodeDirection}(\textit{node \text{m-direction} step})                                                                                                                   & Move node with the given direction and step.                                                                                                                                    \\

\textit{ExtendNode}(\textit{node$_1$ node$_2$})                                                                       & Extend node$_1$ until it hits the node$_2$.                                                                                                                                 \\ 
\textit{ExtendNodeDirection}(\textit{node \text{m-direction}})                                                                       & Extend node in a given direction.                                                                                                                                                    \\ \hline
\end{tabular}
\caption{Example action schemes designed in our DSL with external functions. Whole action descriptions, including preconditions and effects, are available in the supplementary materials.} 
\label{ExampleAction}
\end{table}

\section{Program Synthesis}

We use and improve the PGP($v$) solver~\cite{lei2023novelty} to search in the space of planning programs over training instances of each ARC task. Once the solver returns a program that solves all training instances, we use the test instances to evaluate the solution. The solver core engine is a heuristic search algorithm that starts with an empty program and tries to program an instruction one line at a time until a solution is found.  PGP($v$) uses the notion of \textit{action novelty rank} to scale up the search by pruning a newly generated planning program if its most frequent action repetition is greater than a given bound $v$.

\begin{figure}[!ht]
 
\centering
\includegraphics[scale=0.115]{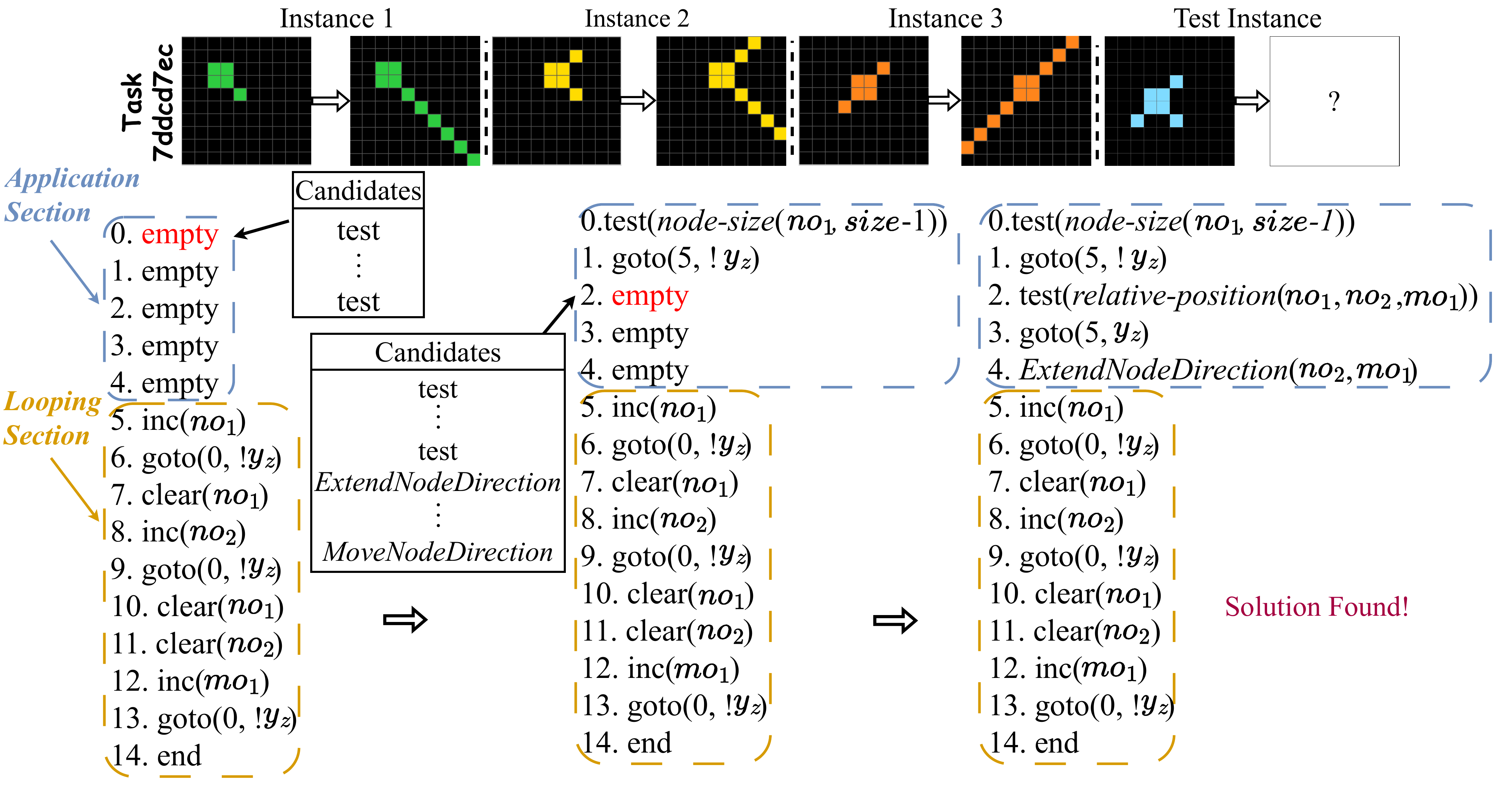}
\caption{An illustration of the planning process with the application section and the looping section. Lines 0 and 1 ensure  $no_1$ indexes the square node, and lines 2 and 3 constrain the $no_2$ to point to the single-pixel node, while $mo_1$ indexes the correct spatial relation between $no_1$ and $no_2$.}
\label{fig8} 
\end{figure}
\subsubsection{Predicate and Argument Constraints}
Predicate constraints limit the allowed arguments of the \texttt{test} action. This action returns the interpretation of a predicate in a program, subsequently used to condition a \texttt{goto} instruction. Predicate constraints are determined before the search starts to ensure only relevant \texttt{test} actions are programmed. We restrict a predicate, describing a node attribute, which can be interpreted by the \texttt{test} action, iff there are two nodes, among all training and test input images, with a distinct value of that attribute. If all image nodes have the same attribute value described by a predicate, then the interpretation of that predicate will not be a helpful condition for a \texttt{goto} instruction, as the interpretation value is always true. For example in the third task of Figure~\ref{fig1}, a valid condition should be the interpretation of the node color predicate rather than node size predicate since all nodes in the input images are of size 1.

Argument constraints make sure that if a node color or size predicate is used in a \texttt{test} action, then the arguments chosen describe attributes that exist in all training and test input images. These constraints prevent overfitting programs to work only on a subset of input instances, increasing the generalizability of the solution programs. For example, conditioning over nodes with size 3 in the second task of Figure~\ref{fig1} would not lead to a valid plan as the node size in the test instance is 2. In this case, other conditions should be used to create a solution moving down every node for one step.

\subsubsection{Structural Restrictions}
Restrictions over the structure of programs are valid strategies to reduce symmetries in the search space~\cite{lei2023novelty}. We adopt structural restrictions by separating a planning program into two sections: the \textit{Application Section} and the \textit{Looping Section}. The  application section can be programmed with planning actions, \texttt{test}, and \texttt{goto} instructions, and the looping section has a sequence of pointer manipulations and \texttt{goto} instructions to ensure the iteration of all possible combinations of pointer values, followed by an $\texttt{end}$ instruction for termination. 

\begin{figure*}[ht]
 
\centering
\includegraphics[scale=0.165]{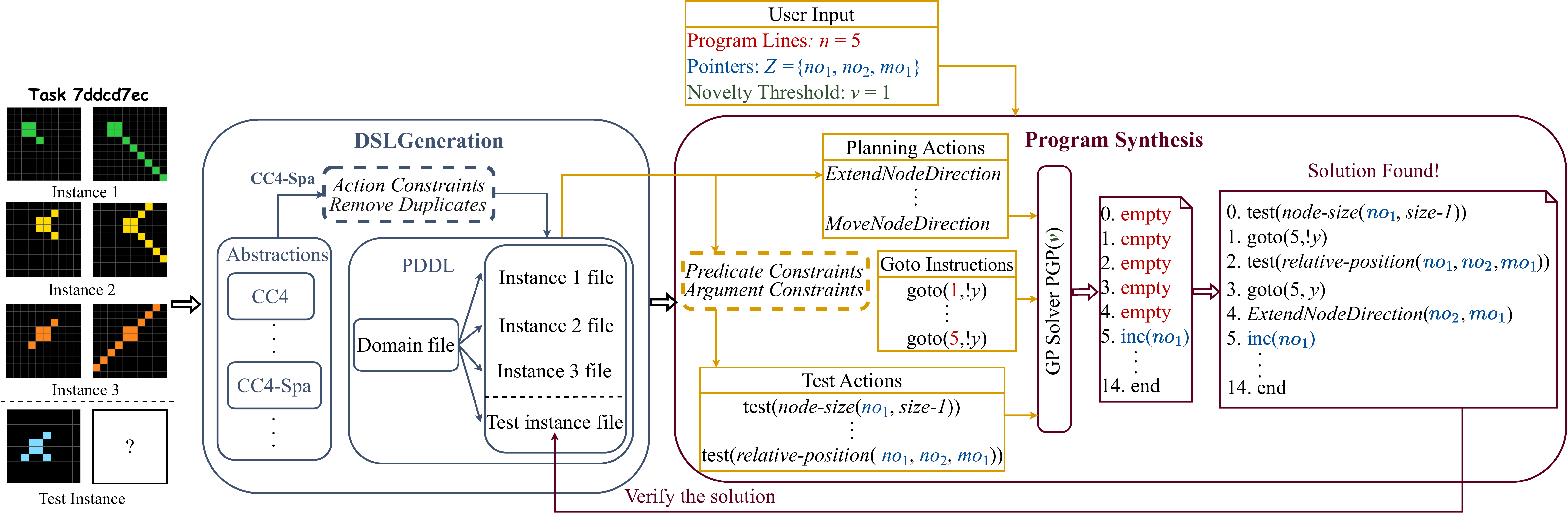}      
\caption{A pipeline sketch of GPAR. CC4  stands for the 4-connected abstraction; CC4-Spa stands for the abstraction that contains complicated movement and extension operations.}
\label{fig9} 
\end{figure*}

We program the looping section before the search starts based on the given pointers. In the application section, the instruction sequence is constrained with the following rules: 1) a \texttt{test} action must be followed by a \texttt{goto} instruction; 2) the first line can only be programmed with a \texttt{test} action; 3) once an action from the DSL is programmed, the subsequent lines must either be programmed with  actions from the DSL or followed by the looping section. To address a scenario where conditions are unnecessary, we include a dummy \texttt{test}(true) action whose interpretation is always true. In GPAR, the number of program lines $n$ refers to the number of lines in the application section rather than in the full program.

Figure \ref{fig8} illustrates the planning process of a planning program with a complex logic: the planning action \textit{ExtendNodeDirection} is executed only when the first tested attribute is false (line 0) and the second tested spatial relation is true (line 2), using three pointers in total. These restrictions come with the caveats of making the solver incomplete. Even if no restrictions are used, existing approaches for ARC are already incomplete, as the expressivity of their DSLs limits the type of solution that can be found.

\subsubsection{Heuristic Function}
Benefiting from the pixel-related predicates, we exploit the pixel information to guide the search. Every time a new program is generated, we execute it, and introduce a heuristic function $h_p$ that goes beyond the goal-count heuristic by counting the number of pixels that differ from the goal state and penalizing further pixels whose values have been changed from the initial state and do not match yet the values in the goal state. This is very similar to the idea in means-ends analysis~\cite{Newell:1963aa}, preferring programs that bring the current state abstraction closer to the goal state. We use $h_p$ to guide the search algorithm used by the solver, and break ties with $h_{ln}$ heuristic, which promotes the application of  action schemes defined in the DSL. Full details of the solver and  $h_{ln}$ can be found in \citet{lei2023novelty}.

\subsubsection{Instantiation over Pointers}
GPAR supports partial instantiation over pointers, where  a subset of parameters in a predicate or action schema are substituted by pointers and others are substituted by objects, such as the planning action shown in Figure \ref{fig3}. This occurs when the number of pointers used to index an object type is less than the number of parameters  specified by that object type.
Partial instantiation allows \texttt{test} actions to fix a specific attribute for looping and branching, and naturally supports parameter bindings~\cite{xu2022graphs} without additional grammar extensions in our DSL.

\begin{table}[]
\captionsetup{font={small}}
\small
\centering
\setlength{\tabcolsep}{3pt}
\renewcommand{\arraystretch}{0.9}
\begin{tabular}{l}
Pointer $\mapsto$ Object Type \\ \hline\hline
$no_1 \mapsto$ \textsc{node}\\
$no_1 \mapsto$ \textsc{node}, $no_2 \mapsto$ \textsc{node} \\
$no_1 \mapsto$ \textsc{node}, $co_1 \mapsto$ \textsc{color} \\
$no_1 \mapsto$ \textsc{node}, $no_2 \mapsto$ \textsc{node}, $co_1 \mapsto$ \textsc{color} \\
$no_1 \mapsto$ \textsc{node}, $co_1 \mapsto$ \textsc{color}, $co_2 \mapsto$ \textsc{color} \\ 
$no_1 \mapsto$ \textsc{node}, $no_2 \mapsto$ \textsc{node}, $mo_1 \mapsto$ \textsc{m-direction} \\
$no_1 \mapsto$ \textsc{node}, $no_2 \mapsto$ \textsc{node}, $no_3 \mapsto$ \textsc{node} \\ \hline
\end{tabular}
\caption{Pointer combinations in GPAR.}
\label{pointers}
\end{table}

\subsection{System Overview}
Figure \ref{fig9} illustrates the pipeline sketch of GPAR, a two-stage system that employs GP to solve ARC tasks. The DSL generation stage encompasses a collection of abstractions with distinct node object, attribute and relation identifications to generate a domain file and associated instance files for each ARC task, where action constraints and duplication removal ensure that only helpful action schemes are included in the domain file, and  unique abstractions are utilized. In the program synthesis stage, ground planning actions and \texttt{test} actions are generated by instantiating action schemes and predicates, described in the domain file, over objects declared in the instance file or pointers given by users,  and \texttt{goto} instructions are generated based on the given program lines. The predicate and argument constraints  increase the likelihood that generated \texttt{test}  actions are useful and goal-oriented. PGP($v$), the GP solver, leverages the user input, program lines $n$, pointers $Z$, and novelty threshold $v$, as parameters to implement the application section and looping section programming. The solution of PGP($v$) is a planning program $\Pi$ that can map the input image, the initial state, to the output image, the goal state, by executing $\Pi$ on the corresponding initial state in each training instance. $\Pi$ is a verified solution if $\Pi$ has been validated as a solution in the test instances.

\begin{table*}[ht]
\centering
\setlength{\tabcolsep}{22.5pt}
\renewcommand{\arraystretch}{1}
\begin{tabular}{lllclc}
\hline
\multicolumn{1}{c}{Model}                                                     & \multicolumn{1}{l}{Task Type} & \multicolumn{2}{c}{ Training Accuracy} & \multicolumn{2}{c}{Testing Accuracy} \\ \hline \hline
\multirow{4}{*}{ARGA}                                                         & movement                      & 18/31             & (58.06\%)           & 17/31            & (54.84\%)           \\
                                                                              & recolor                       & 25/62             & (40.32\%)           & 23/62            & (37.10\%)           \\
                                                                              & augmentation                  & 20/67             & (29.85\%)           & 17/67            & (25.37\%)           \\
                                                                              & all                           & 63/160            & (39.38\%)           & 57/160           & (35.62\%)           \\ \hline
\multirow{4}{*}{\begin{tabular}[c]{@{}l@{}}Kaggle\\ First Place\end{tabular}} & movement                      & \textbf{21}/31             & (\textbf{67.74\%})           & 15/31            & (48.39\%)           \\
                                                                              & recolor                       & 23/62             & (37.10\%)           & 28/62            & (45.16\%)           \\
                                                                              & augmentation                  & \textbf{35}/67             & (\textbf{52.24\%})           & 21/67            & (31.34\%)           \\
                                                                              & all                           & 79/160            & (49.38\%)           & 64/160           & (40.00\%)           \\ \hline
\multirow{4}{*}{GPAR}                                                     & movement                      & 20/31             & (64.52\%)      & \textbf{19}/31            & (\textbf{61.30\%})      \\
                                                                              & recolor                       & \textbf{41}/62             & (\textbf{66.13\%})         & \textbf{39}/62            & (\textbf{62.90\%})         \\
                                                                              & augmentation                  & 25/67             & (37.31\%)           & \textbf{23}/67            & (\textbf{34.33\%})           \\
                                                                              & all                           & \textbf{86}/160            & (\textbf{53.75\%})        & \textbf{81}/160           & (\textbf{50.63\%})\\ \hline         
\end{tabular}
\caption{Performance of ARGA, Kaggle First Place and GPAR over 160 object-centric ARC tasks. Training accuracy is the number of tasks where the solution solves all the training instances. Testing accuracy is the number of tasks where the solution also generates the correct output images for all test instances. Best results are in bold.}
\label{table5}
\end{table*}

\section{Experiments}

As a benchmark, we use the subset of 160 object-centric ARC tasks introduced by \citet{xu2022graphs}. These tasks are further categorized into: 1) \textit{recoloring} tasks which involve changing object colors; 2) \textit{movement} tasks which involve changing object positions; 3) \textit{ augmentation} tasks which involve changing aspects of objects like size or pattern. Figure \ref{fig1} shows example tasks from each class.

\subsection{Parameters}
In GPAR, PGP($v$) takes $n$, $v$, and $Z$ as parameters. The number of program lines $n$ ranges from 3 to 10 where the valid $\Pi$ configuration for $n=3$ is $v=1$ since each instruction included in $\Pi$ with  $n=3$ can only appear once, such as a \texttt{test} action, a \texttt{goto} instruction and a planning action. For $n=4$, reasonable configurations include $v=1$ and $v=2$ since a planning action can appear twice. For $n>4$, the value of $v$ ranges from 1 to 3. All the possible combinations of $Z$ are presented in Table \ref{pointers}, where only object types  \textsc{node}, \textsc{color}, and \textsc{m-direction} are referenced since they are typical specifications of parameters in the design action schemes. The complexity of the search space is proportional to the values of $n$ and $v$. The upper-bound values of $n$ and $v$ ensure the search space is large enough to cover most solutions while still being tractable.

The combination of feasible parameters and a valid DSL is supplied as the input for PGP($v$). For each ARC task, possible combinations are executed in order of increasing complexity, starting from lower values of $n$ and $v$, fewer pointers, and simpler abstractions (e.g., 4-connected are considered before 8-connected abstractions) with a time limit of 1800s for each.  We treat the first encountered $\Pi$ as the solution to generate the test output images for validation. Our approach keeps the search space tractable and ensures we find the simplest solution.

The Kaggle Challenge’s first-place model and ARGA are used as state-of-the-art baselines. All experiments were conducted on a cloud computer with clock speeds of 2.00 GHz Xeon processors. For the Kaggle first-place model and ARGA, the models were executed with a time limit of 1800s per task, and the highest-scored candidate generated by the model is selected as the final solution.
\begin{figure}[ht]
 
\centering
\includegraphics[scale=0.44]{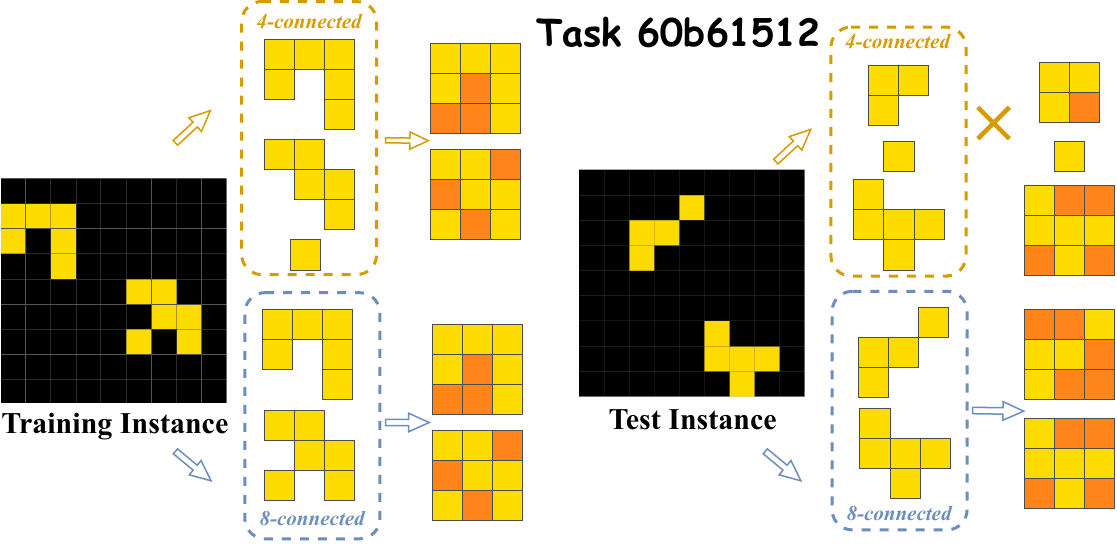}    
\caption{An example task where GPAR generated solution succeeds in the training instance but  fails in the test instance.}
\label{fig13} 
\end{figure}

\subsection{Synthesis of Solutions}

Table \ref{table5} shows the training and testing accuracy of GPAR, ARGA and Kaggle's winner. We score a model as ``correct'' in training if it is able to find a solution that solves all training instances for a given task, and we score that model ``correct'' in testing if its solution also gives the ground truth correct outputs on the test instances. In training, GPAR outperforms the other models in the \textit{recolor} class, and outperforms all other approaches over the test instances. GPAR is the only planner that solves more than half of the tasks, 53.75\% in training and 50.63\% in testing overall. GPAR also shows the great generalization ability, evidenced by the smallest gap between the training and testing accuracy. GPAR  solves 16 tasks exclusively, 13 and 20 tasks overlappingly with Kaggle's winner and ARGA respectively, and 32 tasks commonly by all solvers in testing, illustrated in Figure \ref{veen} in the supplementary materials.

GPAR has a distinct advantage in the \textit{recolor} class, where solutions  are compactly implemented by imperative programs with conditions mainly relying on predicates describing attributes, such as size, shape and color. For the \textit{movement} class, the description of spatial relations remains challenging when dynamic attributes between nodes are needed, such as the center, corner, and area. Meanwhile, some tasks require movement actions defined with large numeric parameters, which is currently not supported well in our DSL. The \textit{augmentation} class involves shape transformations, including rescaling, completion, and analogical replication,  which are difficult to implement in imperative programs based on DSLs. All existing planners struggle with this category, with both training and testing accuracy below 50\%. 

Like previous models, GPAR shows some gap between training and testing, which means that a solution that solves the training set does not generalize to produce the correct results on test instances. Figure \ref{fig13} shows an example where GPAR fails to generalize because both 4-connected and 8-connected abstractions can solve the training instance; however, only the 8-connected abstraction gives the correct solution for the test instance. The correct solution to this task is ambiguous given the training instances.

Of the tasks that GPAR solved in testing, over 50\% require only a novelty threshold of 1 ($v=1$) and just three program lines ($n=3$). The low novelty threshold implies  that most of the tasks can be solved without repeated actions, and the low number of program lines indicates that only a few conditions and/or actions are necessary to produce a solution (44/81 tasks require only one condition). This shows the efficiency of the DSL and  the PGP($v$) solver used by GPAR, which also contributes to its high generalization performance. Figure \ref{Distributions} in the supplementary materials illustrates the distributions of $v$ and $n$ for tasks GAPR solved in testing.

Figure \ref{fig10} in the supplementary materials compares the number of expanded nodes among ARGA, Kaggle First Place, and GPAR, with respect to all solved tasks by each model (``All'') in testing, the subset of tasks solved by all three models (``Common'') in testing, and the subset of tasks solved only by a given model (``Exclusive'') in testing. For GPAR, the count of expanded nodes is accumulated for each execution over the combination of abstraction and reasonable parameters until the first solution $\Pi$ is encountered. GPAR expands more nodes than ARGA, yet fewer than the Kaggle First Place for all solved tasks. If  the number of expanded nodes for the first returned solution is of interest, the performance of GPAR varies widely shown in the leftmost part of Figure \ref{fig11} in the supplementary materials. The ``Common'' tasks solved by all three models generally require a lower number of expanded nodes for GPAR, which also holds true for the first returned solution shown in the middle of Figure \ref{fig11}. Tasks exclusively solved by GPAR tend to require a large number of expanded nodes, surpassing even the count observed in the Kaggle method. However, the number of expanded nodes remains relatively lower for the first returned solution, the rightmost part of Figure \ref{fig11}. The opposite distribution indicates a vast exploration of incorrect abstractions and parameters before discovering the correct solution.

\section{Conclusion}

We leverage an existing solver for generalized planning to synthesize programs with pointers that represent expressive solutions with branches and loops for ARC tasks. We show how the \textit{de facto} language for planning can be used to model object-aware abstractions, resulting in the state-of-the-art performance on the ARC, with greater generalization results. Identifying the most useful  abstractions is still an open problem. In the future, new heuristics can be defined to guide the search of programs through relaxations from the DSL representation, and connections with alternative planning computational models can be explored to improve visual reasoning performance.

\section*{Acknowledgements}

Chao Lei is supported by Melbourne Research Scholarship established by The University of Melbourne.

This research was supported by use of The University of Melbourne Research Cloud, a collaborative Australian research platform supported by the National Collaborative Research Infrastructure Strategy (NCRIS).

\bibliography{aaai24}

\clearpage
\onecolumn
\appendix
\appendixpage

\begin{table*}[!ht]
\captionsetup{}
\centering
\small
\setlength{\tabcolsep}{5pt}
\renewcommand{\arraystretch}{1.3}
\begin{tabular}{p{0.26\columnwidth}p{0.21\columnwidth}p{0.25\columnwidth}p{0.2\columnwidth}}

Abstractions (Abbreviations)  & Node Definitions              & Included Predicates       & Included Action Schemes \\ \hline\hline
\textit{4-connected}$^*$ (CC4)       & \multicolumn{1}{c}{$\circ$}                                                           & \textit{node-color, node-size, node-shape, contain-node, partially-contain-node, node-spatial, node-diagonal, odd, even, size-max, size-min, color-max, color-min, background, contain-pixel, pixel-color.}                                         & \textit{UpdateColor, MoveNode1, ExtendNode, AddBorder, MirrorNode, InsertNode, HollowNode, FillNode.} \\

\textit{4-connected black}$^*$ (CC4-B)         & 4-connected components as nodes excluding the black pixels.                                                    & \multicolumn{1}{c}{ -}                                                                                                                                                                                                                                                                    & \multicolumn{1}{c}{-}                                                                                           \\
\textit{4-connected all$^*$}(CC4-All)         & 4-connected components as nodes in both background and non-background regions.                                         &     \multicolumn{1}{c}{- }                                                                                                                                                                                                                                                                   &        \multicolumn{1}{c}{-}                                                                                   \\
\textit{4-connected spatial}$^*$  (CC4-Spa) & 
Same as CC4. & \textit{node-color, node-size, node-shape, contain-node, partially-contain-node, node-spatial, relative-position, node-diagonal, background, contain-pixel, pixel-color.}                                                                                            & \textit{UpdateColor, MoveNode2, MoveNodeDirection1, MoveNodeDirection2, ExtendNodeDriection.}    \\
\textit{4-connected congruent}$^*$ (CC4-Con) & 
Same as CC4.
& \textit{node-color, node-size, contain-node, partially-contain-node, node-spatial, same-color, congruent, color-max, color-min,  background, contain-pixel, pixel-color.}                                                                                       & \textit{UpdateColor, SwapColor, CopyColor.}                                            \\

\textit{multi-color}$^*$ (MC)         & \multicolumn{1}{c}{$\circ$  }                              & \textit{node-size, node-shape, contain-node, partially-contain-node, node-spatial, node-diagonal, size-max, size-min, even, odd, background, contain-pixel, pixel-color.}                                                                            & \textit{MoveNode1, ExtendNode, AddBorder, MirrorNode, InsertNode, HollowNode, FillNode.}             \\

\textit{max-rectangle} (Re)           & \multicolumn{1}{c}{$\circ$  }                                                &          \multicolumn{1}{c}{-}                                                                                                                                                                                                                                                             & \multicolumn{1}{c}{-}  \\ 
\textit{background-rectangle} (BG-Re)    &  Maximum rectangle identifications within 4-connected components in background regions.                                                &              \multicolumn{1}{c}{-}                                                                                                                                                                                                                                                         &             \textit{UpdateColor.}                                                                                \\

\textit{same-color} (SC)     & \multicolumn{1}{c}{$\circ$  }                                                                          &                                  \multicolumn{1}{c}{-}                                                                                                                                                                                                                                      & \textit{UpdateColor, SwapColor, CopyColor.} \\   

\textit{pixels}            & \multicolumn{1}{c}{$\circ$  }                                                                                 & \textit{node-color, background, color-most, color-least, denosing-color, right-diagonal, left-diagonal, up-border, down-border, left-border, right-border, vertical-middle, horizontal-middle, contain-pixel, pixel-color.} & \textit{UpdateColor.}                                                                             \\
\textit{image}             & \multicolumn{1}{c}{$\circ$  }                                                                                & \textit{contain-pixel, pixel-color.}                                                                                                                                                                                                                                           & \textit{RotateNode, FlipNode.}                                                                        \\ \textit{horizontal}         & \multicolumn{1}{c}{$\circ$  }                           &           \multicolumn{1}{c}{-}                                                                                                                                                                                                                                                            &          \multicolumn{1}{c}{-}                                                                                 \\
\textit{vertical}         &  \multicolumn{1}{c}{$\circ$  }                                                                             &         \multicolumn{1}{c}{-}                                                                                                                                                                                                                                                               &         \multicolumn{1}{c}{-}                                                                                 \\\hline                                                      
\end{tabular}

\caption{Abstractions with related node definitions, predicates, and action schemes, represented by names. The superscript ``$^*$'' denotes that the 8-connected version is considered; the symbol ``$\circ$'' denotes that the node definition has been given in the manuscript;  the symbol ``-''  stands for the same predicates or action schemes allocated as the CC4 version. CC4-Spa and CC4-Con are two abstractions that enable complicated movement and extension operations, and congruent node operations respectively. The background region is connected pixels, with the background color, which reach one of the image boundaries, where the background color is black if black exists; otherwise, it is the most frequent color in the image. }
\label{abstractions}
\end{table*}

\begin{table*}[!ht]
\captionsetup{}
\centering
\small
\setlength{\tabcolsep}{5pt}
\renewcommand{\arraystretch}{1.4}
\begin{tabular}{p{0.33\columnwidth}p{0.28\columnwidth}p{0.33\columnwidth}}

Action Schemes(? Parameters)                                    & Preconditions                                                                                                                                                               & Effects                                                                                                                                                              \\ \hline \hline
\textit{UpdateColor}(\textit{node color$_1$ color$_2$}) & \textit{node-color}(\textit{node color$_1$})$=True$                                                                                                                           & Change the node color from color$_1$ to color$_2$.                                                                                                                          \\
\textit{CopyColor}(\textit{node$_1$ node$_2$})          & \textit{node$_1$} $\neq$ \textit{node$_2$}                                                                                                      & Copy the color of node$_1$ to node$_2$.                                                                                                                                         \\ 
\textit{SwapColor}(\textit{node$_1$ node$_2$})          & \textit{node$_1$} $\neq$ \textit{node$_2$}                                                                                                       & Swap colors of node$_1$  and node$_2$.                                                                                                                                        \\ 
\textit{MoveNode1}(\textit{node$_1$ node$_2$})          & $\exists$
$dr \in $ \textit{M-DIRECTION}, 
\textit{node-spatial}(\textit{node$_1$ node$_2$ dr})$=True$                                                                                                               & Move node$_1$ to the border of node$_2$.                                                                                                                                    \\ 
\textit{MoveNode2}(\textit{node$_1$ node$_2$})          &  $\exists$
$dr \in $ \textit{M-DIRECTION},  
\textit{node-spatial}(\textit{node$_1$ node$_2$ dr})$=True$                                                                                  & Recursively move nodes located between node$_1$ and node$_2$  to the border of node$_2$ until node$_1$ is moved. \\ 


\textit{MoveNodeDirection1}(\textit{node m-direction})   & \multicolumn{1}{c}{-}                                                                                                                          & Recursively move nodes located between the given node and the image boundary in the given direction until the given node is moved.     \\


\textit{MoveNodeDirection2}(\textit{node m-direction step})     & \multicolumn{1}{c}{-}                                                                                                                            & Move the node in the given direction with the given step until it reaches the image boundary.                                                             \\  
\textit{ExtendNode}(\textit{node$_1$ node$_2$})         &  $\exists$
$dr \in $ \textit{M-DIRECTION},
\textit{node-spatial}(\textit{node$_1$ node$_2$ dr})$=True$                                                                                                              & Extend node$_1$ until it hits the node$_2$.                                                                                                                                 \\ 
\textit{ExtendNodeDirection}(\textit{node m-direction})         & \multicolumn{1}{c}{-}                                                                                                                                 & Extend the node in the given direction until it hits other nodes or the image boundary.                                                     \\ 
\textit{RotateNode}(\textit{node degree})         &  \textit{node$^*$} $\in |image| \land$ \textit{node$^*$} $\cap$ \textit{Nodes}$= \emptyset$                                                                           & Rotate the node with the given degree.                                                                                                                                    \\ 
\textit{HollowNode}(\textit{node color})          & \textit{node-shape}(\textit{node square})$=True$ $\lor$   \textit{node-shape}(\textit{node rectangle})$=True$               & Hollow the node with the given color.                                                                                                                                  \\ 
\textit{AddBorder}(\textit{node color})           &  \textit{node$^*$} $\in |image| \land$ \textit{node$^*$} $\cap$ \textit{Nodes}$= \emptyset$   & Add a closed border to the given node with the given color.                                                                                                                                  \\ 
\textit{MirrorNode}(\textit{node$_1$  node$_2$})  &  \textit{node$_1$} $\neq$ \textit{node$_2$} $\land$ \textit{node$^*$} $\in |image| \land$ \textit{node$^*$} $\cap$ \textit{Nodes}$= \emptyset$                                                                               & Mirror node$_1$ based on node$_2$.                                                                                                                                         \\ 
\textit{FlipNode}(\textit{node f-direction})        & \textit{node$^*$} $\in |image| \land$ \textit{node$^*$} $\cap$ \textit{Nodes}$= \emptyset$                                                                               & Flip the node with given direction.                                                                                                                                   \\ 
\textit{InsertNode}(\textit{node$_1$ node$_2$})         &  \textit{node$_1$} $\neq$ \textit{node$_2$} $\land$ \textit{node$^*$} $\in |image| \land$ \textit{node$^*$} $\cap$ \textit{Nodes}$= \emptyset$                                                                        & Delete node$_2$ and insert node$_1$ to  the position of node$_2$.                                                                \\ 
\textit{FillNode}(\textit{node color})            & \textit{node-shape}(\textit{node unknown})$=True$$\land$ \textit{node$^*$} $\in |image| \land$ \textit{node$^*$} $\cap$ \textit{Nodes}$= \emptyset$                                                                                                                              & Fill a node to be the rectangle  with the given color.                                                                                                                                  \\ \hline
\end{tabular}

\caption{Action schemes designed for DSL implemented by external functions with respect to preconditions and effects.  The symbol ``-''  denotes no precondition; \textit{node$^*$} represents the resulting node after the transformation;  \textit{node$^*$} $\in |image|$ means  that \textit{node$^*$} is inside of the image boundary; \textit{node$^*$} $\cap$ \textit{Nodes}$= \emptyset$  stands for \textit{node$^*$} is not collision with existing nodes except itself.}
\label{actions}

\end{table*}

\begin{table}[!ht]
\captionsetup{}
\centering
\small
\setlength{\tabcolsep}{5pt}
\renewcommand{\arraystretch}{1.3}
\begin{tabular}{p{0.13\columnwidth}p{0.4\columnwidth}p{0.4\columnwidth}} 
Constraints                      & Conditions                                  & Pruned Action Schemes                                                                                                                                              \\ \hline \hline
positionUnchanged & Node does not change position after update. & \textit{MoveNodeDirection1, MoveNodeDirection2, MoveNode1, MoveNode2, ExtendNodeDirection, ExtendNode, AddBorder, InsertNode, FillNode, MirrorNode, RotateNode, FlipNode.} \\
colorUnchanged    & Node does not change color after update.    & \textit{UpdateColor.}                                                                                                                             \\
sizeUnchanged    & Node does not change size after update.  & \textit{InsertNode, ExtendNode, ExtendNodeDirection, AddBorder, FillNode, HollowNode.
}
\\
fillConstraint    & None of the nodes with shape unknown in input images.  & \textit{ FillNode.} 

\\

hollowConstraint    & None of the nodes with shape square or rectangle in  input images.  & \textit{ HollowNode}.   

\\

insertConstraint    & No consistent pattern (nodes with the
same color, size, and shape) exists among input images.  & \textit{InsertNode}.   

\\ \hline

\end{tabular}
                      
\caption{Example constraints and related pruned action schemes, represented by names.}
\label{tableConstrain}
\end{table}

\twocolumn

\begin{figure}[!ht]
 
\centering
\includegraphics[scale=0.3]{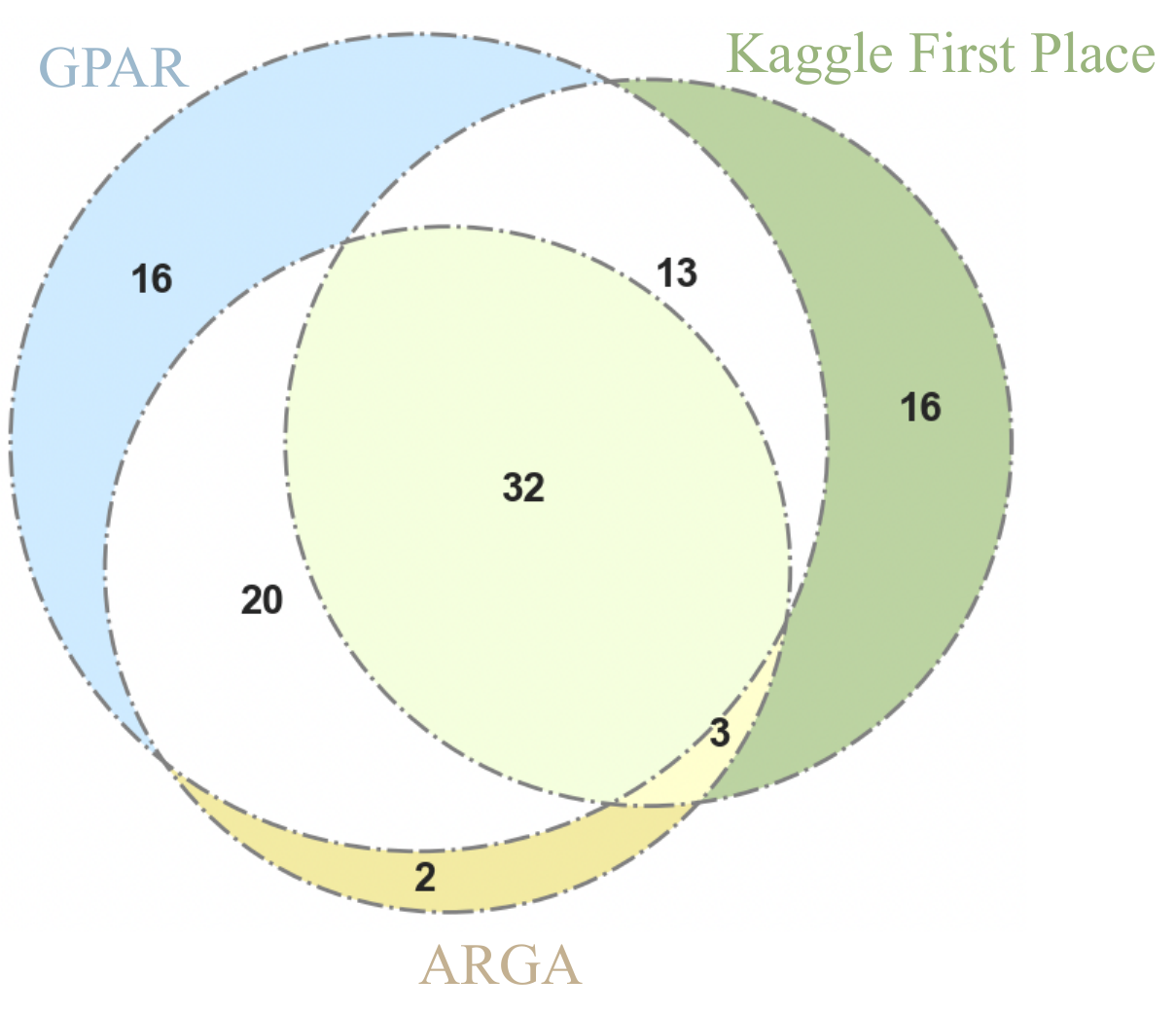}  

\caption{ The Venn diagram of the number of solved tasks by GPAR, Kaggle First Place, and ARGA in testing.}
\label{veen} 
\end{figure}

\begin{figure}[!h]
 
\centering
\includegraphics[scale=0.18]{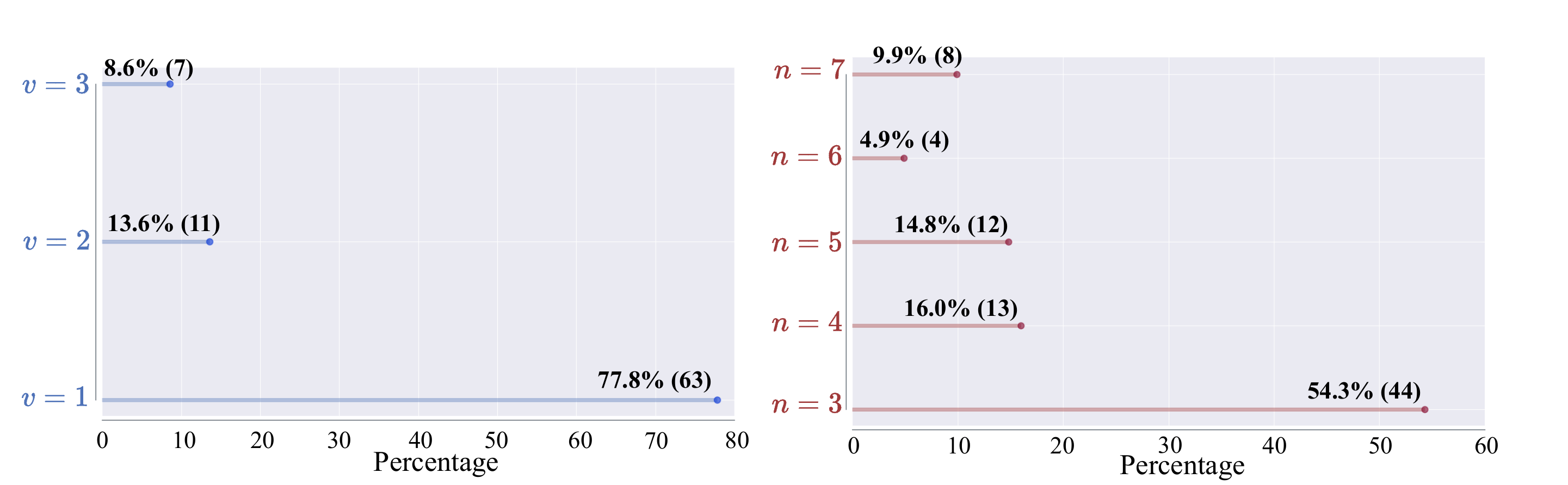}
\caption{Distributions of the values of  novelty threshold $v$, and program lines $n$, for GPAR solved tasks in testing.}
\label{Distributions} 
\end{figure}

\begin{figure}[ht]
 
\centering
\includegraphics[scale=0.48]{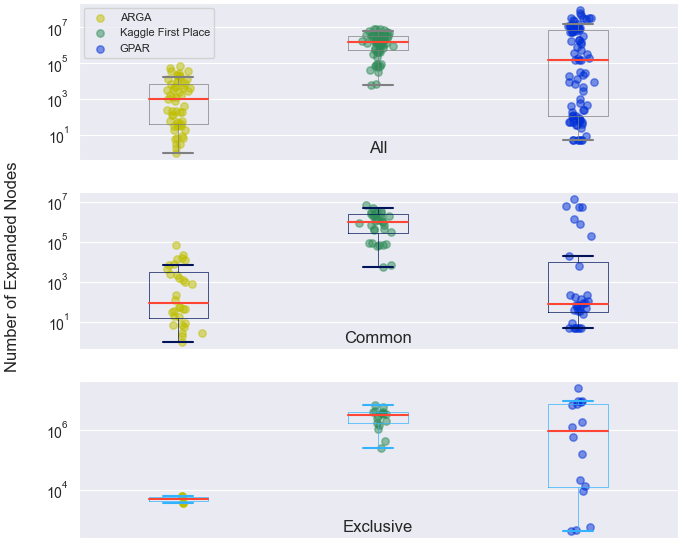}        \vspace{-0.1cm}
\caption{Comparisons of the number of expanded nodes among ARGA, Kaggle First Place, and GPAR, shown in the upper figure for all solved tasks, the middle figure for commonly solved 32 tasks, and the lower figure for exclusively solved tasks in testing. }
\label{fig10} 
\end{figure}

\begin{figure}[ht]
 
\centering
\includegraphics[scale=0.48]{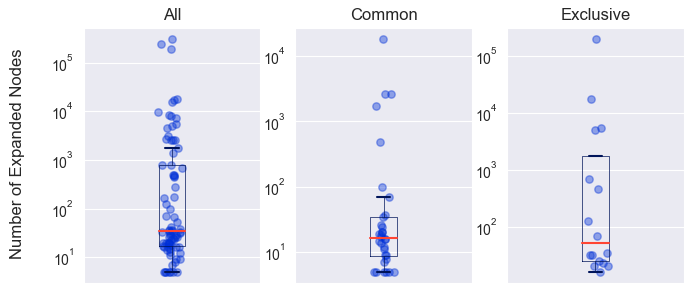}   
\caption{The number of expanded nodes with respect to the first returned solution in GPAR for all solved (left), commonly solved (middle), and exclusively solved tasks (right) in testing.}
\label{fig11} 
\end{figure}

\end{document}